\documentclass{ieeetmlcn}
\usepackage{cite}
\usepackage{amsmath,amssymb,amsfonts}
\usepackage{mathtools}
\usepackage{algorithmic}
\usepackage{graphicx,color}
\usepackage{textcomp}
\usepackage{xcolor}
\usepackage{hyperref}
\hypersetup{hidelinks=true}
\usepackage{algorithm,algorithmic}
\usepackage{subfig}
\usepackage{booktabs}
\usepackage{mathrsfs}
\usepackage{comment}
\usepackage{balance}

\makeatletter
\@ifundefined{proposition}{\newtheorem{proposition}{Proposition}}{}
\makeatother

\def\BibTeX{{\rm B\kern-.05em{\sc i\kern-.025em b}\kern-.08em
    T\kern-.1667em\lower.7ex\hbox{E}\kern-.125emX}}
\AtBeginDocument{\definecolor{tmlcncolor}{cmyk}{0.93,0.59,0.15,0.02}\definecolor{NavyBlue}{RGB}{0,86,125}}

\def\OJlogo{}
\def\seclogo{}

\makeatletter
\AtBeginDocument{%
  \def\@evenfoot{\hbox to \textwidth{{\rffont\thepage}\hfill}}%
  \def\@oddfoot{\hbox to \textwidth{\hfill{\rffont\thepage}}}%
  \def\@evenhead{\vbox{\hsize\textwidth\vbox to 0pt{\hsize\textwidth\vspace*{7.7pt}\rfxfont\raggedright\rightmark\hfill\par}\par\vspace*{16pt}\hbox to \textwidth{\vrule width\textwidth height.3pt depth0pt}}}%
}
\makeatother

\makeatletter
\def\@maketitle{\newpage
  \bgroup%
    \vspace*{-10.5pt}%
    \ifx\@doiinfo\@empty\else\vskip3pt{\doi@font Digital Object Identifier 10.1109/\@doiinfo\par}\fi%
     \vskip14.5pt{\titlefont{\textcolor{tmlcncolor}{\@title}}\par\vskip8.5pt}%
    {\authorfont\@author\par}\vskip4pt{\afffont\@affil}\par\@IEEEspecialpapernotice\vskip5pt%
    \ifx\@corresp\@empty\else{\corfont\@corresp\par}\vskip5.5pt\fi%
    \ifx\@authornote\@empty\else{\afffont\@authornote\par}\vskip27.5pt\fi%
    \centerline{\rule{36pc}{.5pt}}\par\vskip11.5pt%
    \ifvoid\abstractbox\else{\reset@font\box\abstractbox}\fi\par\vskip11.5pt%
    \ifvoid\keybox\else{\reset@font\box\keybox}\fi\par\vskip-32pt%
    \centerline{\rule{36pc}{.5pt}}%
    \par\addvspace{33.5pt}\egroup%
    \let\maketitle\relax
}%
\makeatother

\def\authorrefmark#1{\ensuremath{^{\textbf{#1}}}}

\begin{document}

\markboth{}{Farzaneh and Simeone}

\title{Online Conformal Anomaly Detection with Prediction-Powered Data Acquisition}

\author{Amirmohammad Farzaneh\authorrefmark{1} and Osvaldo Simeone\authorrefmark{1}}
\affil{Institute for Intelligent Networked Systems (INSI), Northeastern University London, London, UK\authorrefmark{1}}
\corresp{Corresponding author: Amirmohammad Farzaneh (email: a.farzaneh@nulondon.ac.uk).}

\begin{abstract}
Online anomaly detection is essential in fields such as cybersecurity, healthcare, industrial monitoring, and telecommunications, where promptly identifying deviations from expected behavior can avert critical failures or security breaches. While numerous anomaly scoring methods based on supervised or unsupervised learning have been proposed, the only existing approach capable of providing assumption-free guarantees on the false discovery rate (FDR) rely on a continuous stream of real-world calibration data. To address this limitation, we introduce context-aware prediction-powered conformal online anomaly detection (C-PP-COAD), a novel principled framework that strategically leverages synthetic calibration data to mitigate data scarcity, while adaptively integrating real data based on contextual information. C-PP-COAD wraps around any existing anomaly detection method, leveraging any given anomaly score to construct active conformal p-value statistics. These statistics support online testing with formal FDR control, maintaining rigorous and reliable anomaly detection performance over time. Experiments conducted on both synthetic and real-world datasets, including thyroid dysfunction detection, O-RAN conflict detection, 5G network intrusion detection, and O-RAN UE throughput degradation detection, demonstrate that C-PP-COAD significantly reduces dependency on real calibration data without compromising guaranteed FDR control.
\end{abstract}


\maketitle

\section{INTRODUCTION}
\label{sec:intro}

\begin{figure}[t!]
    \centering
    \subfloat[]{\includegraphics[width=\columnwidth]{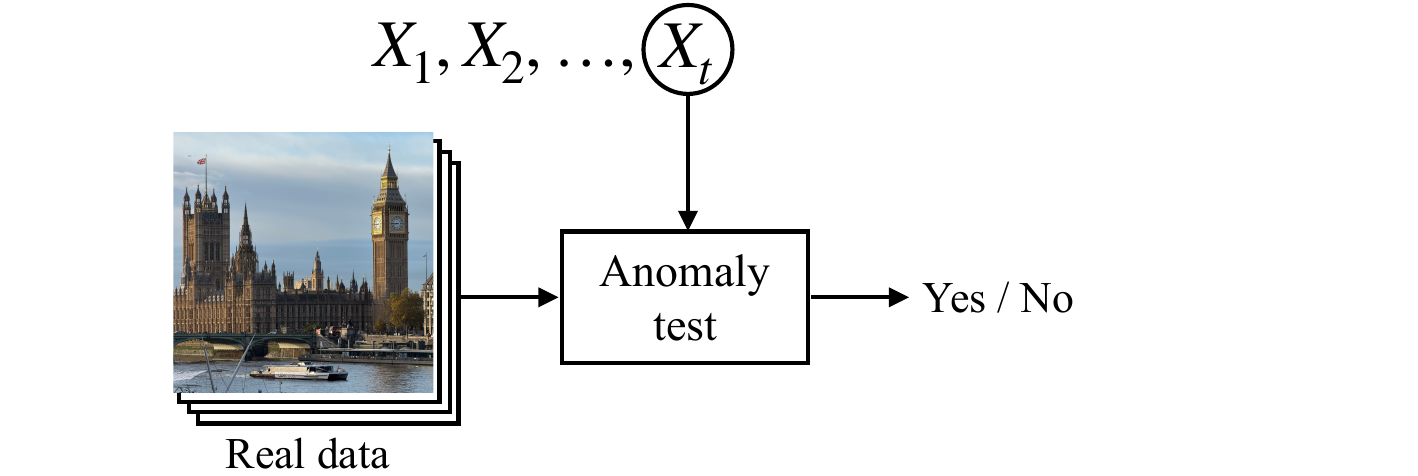}}
    \\
    \subfloat[]{\includegraphics[width=\columnwidth]{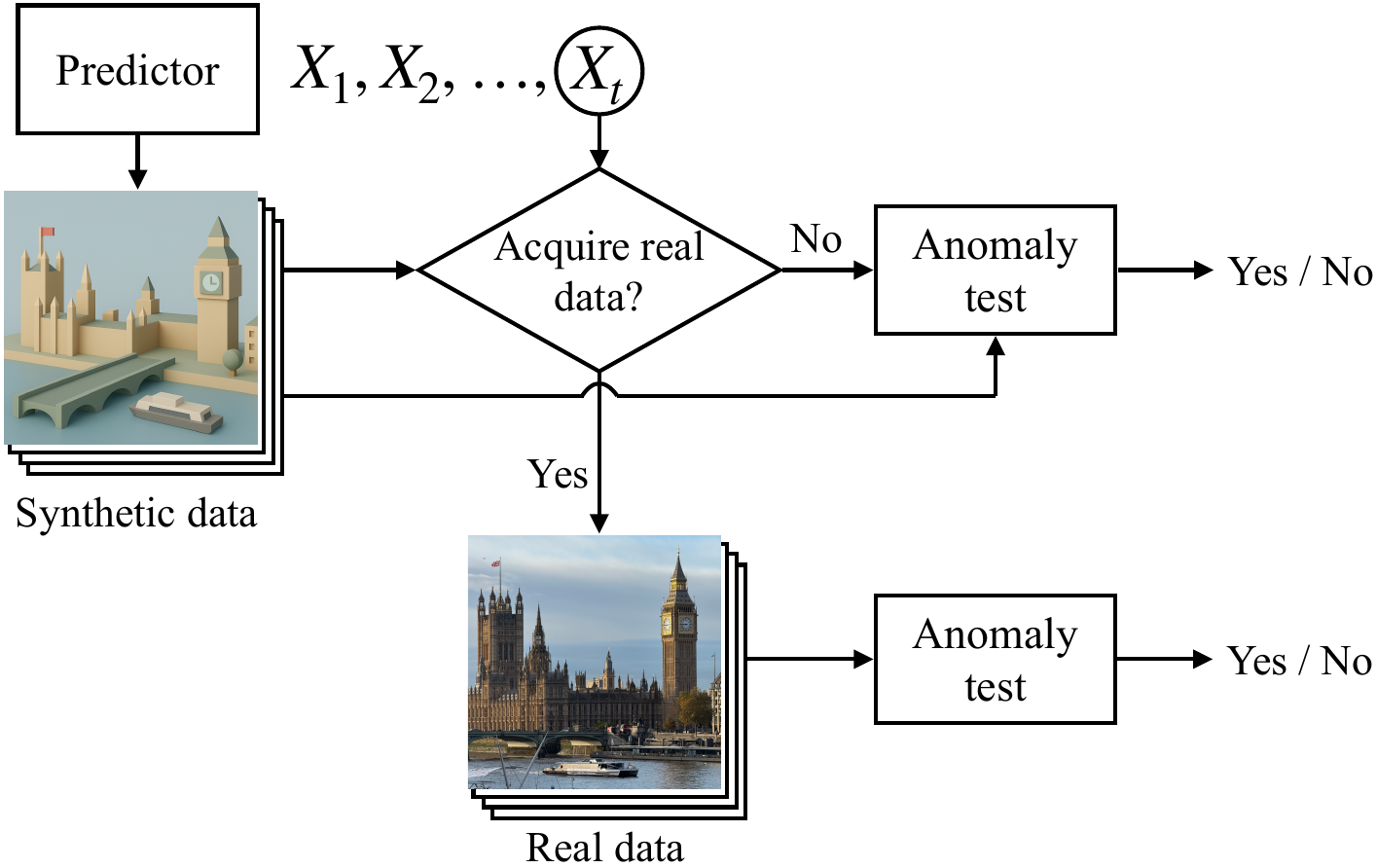}}
    \caption{(a) Conventional anomaly detection approaches with FDR control -- referred to as \textit{conformal online anomaly detection} (COAD) -- require a continuous stream of fresh nominal data to recalibrate the scoring function used in the anomaly test. (b) The proposed approach, \textit{context-aware prediction-powered conformal online anomaly detection} (C-PP-COAD), adaptively chooses between real and synthetic data using contextual information, so as to improve data efficiency while maintaining statistical reliability.}
    \label{fig:tradeoff_figure}
\end{figure}

\IEEEPARstart{O}{nline} anomaly detection underpins safety- and security-critical monitoring in domains such as cybersecurity, healthcare, finance, telecommunications, and industrial systems \cite{sommer2010outside,west2016intelligent,hauskrecht2013outlier,parwez2017big,mahrez2023benchmarking}. At each time $t$, a detector assigns an \emph{anomaly score} to an incoming observation (possibly conditioned on context) and decides whether to flag it as anomalous. While many supervised, semi-supervised, and unsupervised scoring functions have been proposed, e.g., autoencoder reconstruction errors \cite{zhou2017anomaly,nguyen2024variational}, clustering distances \cite{munz2007traffic}, isolation forests \cite{liu2008isolation}, autoregressive prediction errors \cite{hundman2018detecting}, one-class SVMs \cite{scholkopf2001estimating}, deep SVDD \cite{ruff2018deep}, and binary classifiers \cite{goldstein2016comparative}, their operational deployment often requires \emph{reliability guarantees} on false alarms.

We focus on \emph{online} settings in which decisions are made sequentially as data arrive. Recent work on \emph{conformal online anomaly detection} (COAD) obtains finite-sample, distribution-free control of a time-averaged false discovery rate (FDR) by converting scores into conformal p-values using \emph{fresh nominal calibration data} at each time step and applying online FDR control rules \cite{rebjock2021online,lee2025full}. The key bottleneck is that COAD requires freshly acquired real calibration data at every time step. In telecommunications, this is particularly limiting: newly deployed network slices, rare traffic contexts, or privacy-constrained environments often prevent continuous collection of labeled nominal data \cite{edozie2025artificial}.

Modern 5G networks and O-RAN deployments routinely maintain \emph{digital twins} (DTs), high-fidelity virtual models capable of generating context-specific nominal samples on demand \cite{jones2020characterising,nguyen2021digital,pegurri2025toward,ruah2024calibrating}. A natural first step is to replace real calibration data with DT-generated synthetic samples entirely. This \emph{prediction-only} scheme (PO-COAD) incurs zero real data acquisition cost. However, the inevitable distribution mismatch between the DT and the true nominal distribution makes the approach statistically invalid \cite{bashari2025synthetic,bashari2025statistical}. As we show empirically in Sec.~\ref{sec:experiments}, PO-COAD fails to control the FDR, even when the DT is of high quality.

To restore FDR validity without abandoning the efficiency of synthetic data, we build on the \emph{active p-value} framework of reference~\cite{xu2025active}. The key idea is to use statistics obtained from synthetic data as a \emph{gating signal} rather than a direct test statistic: when the synthetic statistic suggests nominal conditions, real data is unlikely to change the decision and can safely be skipped. Conversely, when synthetic data suggests an anomaly, real data is queried with high probability and a valid statistic is computed. A carefully designed randomized combination of synthetic and real statistics yields a valid statistic under the null hypothesis. The resulting novel scheme, referred to as \emph{prediction-powered COAD} (PP-COAD), preserves FDR control at every time step.

The full proposed framework, \emph{context-aware prediction-powered COAD} (C-PP-COAD), applies a context-specific combination of real and synthetic statistics, allowing a tailored design, calibrated to the DT's fidelity in each operating regime. Better DT quality in a given context yields a higher weight to synthetic data and fewer real data queries.

A comprehensive survey of related work and a detailed positioning of C-PP-COAD against existing methods are provided in Sec.~\ref{sec:related}.

\paragraph{Main contributions.}
We introduce C-PP-COAD, a novel online anomaly detection framework that wraps around any pre-trained anomaly scoring function and is agnostic to the type of DT used (see Fig.~\ref{fig:tradeoff_figure}). The main contributions are as follows:
\begin{itemize}
    \item \textbf{Prediction-powered COAD (PP-COAD):} We first present PP-COAD, which combines proxy conformal p-values from a digital twin with real conformal p-values via the active p-value framework of \cite{xu2025active}. By using the proxy as a gating signal, PP-COAD selectively queries real calibration data only when the proxy is uncertain, substantially reducing real data cost while retaining statistical validity and FDR control at all times.
    \item \textbf{Context-aware PP-COAD (C-PP-COAD):} We further generalize PP-COAD by introducing a context-dependent acquisition parameter, calibrated to the DT's per-context fidelity. C-PP-COAD reduces real data cost in contexts where the DT is reliable and improves detection power in contexts where it is informative.
    \item \textbf{Statistical validity:} We prove that C-PP-COAD guarantees sFDR control at the target level $\alpha$ for any anomaly score function and any sequence of context variables (Proposition~\ref{prop:prop}).
    \item \textbf{Experiments:} We evaluate C-PP-COAD on four datasets: thyroid dysfunction detection \cite{asuncion2007uci}, O-RAN conflict detection \cite{shami2025ran}, 5G network intrusion detection \cite{samarakoon2022nidd}, and O-RAN UE throughput degradation detection \cite{polese2022coloran}. C-PP-COAD consistently reduces real data acquisition while maintaining FDR control and competitive detection power.
\end{itemize}

\paragraph{Organization.}
Sec.~\ref{sec:related} surveys related work in anomaly detection, with a specific focus on telecom applications \cite{chandola2009anomaly,edozie2025artificial,mahrez2023benchmarking}, conformal anomaly detection \cite{bates2023testing,rebjock2021online,lee2025full}, and digital twins \cite{jones2020characterising,nguyen2021digital,ruah2024calibrating}, positioning C-PP-COAD against the state of the art. Sec.~\ref{sec:problem} formulates the context-aware online anomaly detection problem and objectives. Sec.~\ref{sec:methodology} presents C-PP-COAD and its theoretical guarantees. Sec.~\ref{sec:experiments} provides experimental results, and Sec.~\ref{sec:conclusion} concludes. Appendices provide additional background, methodological details, extensions, and experimental results.

\section{RELATED WORK}
\label{sec:related}

In this section, we survey related work across three areas: anomaly detection in telecommunications, conformal prediction and online FDR control, and digital twins for network monitoring. We then position C-PP-COAD against the state of the art.

\subsection{ANOMALY DETECTION IN TELECOMMUNICATIONS}

Anomaly detection plays a key role in fields as diverse as cybersecurity \cite{sommer2010outside}, healthcare \cite{hauskrecht2013outlier}, finance \cite{west2016intelligent}, and industrial monitoring \cite{parwez2017big}. In telecommunication networks, anomaly detection has been extensively studied, driven by the need to detect faults, security breaches, and performance degradations in increasingly complex network infrastructures \cite{chandola2009anomaly,edozie2025artificial}. Existing approaches span three main categories: threshold-based methods, machine learning (ML)-based methods, and online/adaptive methods.

\textbf{Threshold-based methods:} Classical approaches rely on manually set thresholds or statistical process control techniques, such as exponentially weighted moving average (EWMA) charts and seasonal forecasting models, to detect deviations in key performance indicator (KPI) time series \cite{wang2018anomaly}. While computationally lightweight, these methods require domain expertise to configure, are sensitive to distribution shifts and operating regime changes, and provide no formal guarantees on false alarm rates \cite{mahrez2023benchmarking}.

\textbf{Machine learning methods:} The growth of large-scale network datasets has enabled data-driven detection approaches. Autoencoder-based methods detect anomalies via reconstruction errors on KPI vectors \cite{zhou2017anomaly,nguyen2024variational}, and LSTM-based predictors are widely used for temporal KPI forecasting, with anomalies declared when prediction errors exceed empirical thresholds \cite{hundman2018detecting}. Isolation forests \cite{liu2008isolation} and one-class SVMs \cite{scholkopf2001estimating} have been applied to network intrusion detection, traffic anomaly detection \cite{munz2007traffic}, and user activity monitoring in mobile wireless networks \cite{parwez2017big}. More recently, graph neural network (GNN)-based methods have been proposed for xApp conflict detection in open radio access networks (O-RAN), capturing structured dependencies among network elements \cite{shami2025ran,adamczyk2023conflict}. A comprehensive benchmarking study for anomaly detection in O-RAN handover optimization highlights the gap between empirical performance and formal reliability guarantees in this context \cite{mahrez2023benchmarking}. A recent survey of artificial intelligence advances in telecom anomaly detection \cite{edozie2025artificial} confirms that no existing telecom-specific method provides formal control of false discovery rates.

All of these ML-based methods share a fundamental limitation: they rely on ad hoc thresholding, such as empirical quantiles of training-set scores or cross-validation, to declare anomalies, with no formal control of false discovery rates. Setting thresholds by heuristics means the resulting false alarm rate is uncontrolled and may vary substantially across deployment conditions.

\textbf{Online and adaptive methods:} Real-time network monitoring requires methods that adapt to non-stationary KPI distributions caused by load variations, network reconfigurations, or time-of-day patterns \cite{wette2024oml,ahmad2017unsupervised}. Online machine learning methods maintain rolling estimates of the nominal distribution and flag deviations as they arise. However, despite their adaptability, existing online methods for telecommunications, including those explicitly designed for LTE/5G environments \cite{nugroho2023performance,li2019anomaly}, do not address the problem of controlling false anomaly detections over time.

\subsection{CONFORMAL PREDICTION AND ONLINE FDR CONTROL}

Conformal prediction \cite{angelopoulos2023conformal} provides distribution-free, finite-sample coverage guarantees for prediction sets and has been extended to hypothesis testing, including outlier and anomaly detection \cite{bates2023testing,xu2021conformal,bashari2025robust}. Conformal p-values are constructed from exchangeable calibration data and satisfy the superuniformity condition $\Pr[P \leq u \mid \mathcal{H}_0] \leq u$ under the null hypothesis, enabling principled anomaly testing without distributional assumptions.

In the online setting, sequential anomaly detections correspond to a stream of hypothesis tests, and controlling the proportion of false detections over time requires online FDR control procedures. Methods such as LORD \cite{ramdas2017online}, SAFFRON \cite{ramdas2018saffron}, and ADDIS \cite{tian2019addis} maintain FDR control at every time step while allowing adaptive rejection thresholds. COAD \cite{rebjock2021online} unifies conformal p-values with the LORD procedure to yield the first online anomaly detection framework with finite-sample, distribution-free FDR guarantees. More recently, e-value-based approaches have been proposed to improve detection power in this setting \cite{lee2025full}. \emph{Model-based} sequential change detection \cite{lai1995sequential,tartakovsky2014sequential} provides guarantees under parametric assumptions but is less suited to localized, sample-wise anomaly decisions and typically yields asymptotic guarantees.

None of these methods, however, address the practical constraint of limited calibration data. COAD and its extensions assume that a fresh batch of calibration samples from the nominal distribution is available at every time step. In telecommunications, this is often infeasible: newly deployed network slices, rare operating contexts, or privacy-constrained environments limit the availability of real nominal data. C-PP-COAD addresses this limitation directly.

\subsection{DIGITAL TWINS FOR NETWORK MONITORING}

Digital twins (DTs) are high-fidelity virtual replicas of physical systems, capable of simulating their behavior under arbitrary conditions \cite{jones2020characterising}. In telecommunications, DTs have been proposed for network planning, resource optimization, and predictive maintenance in 5G and beyond \cite{nguyen2021digital,khan2022digital}. Ray-tracing-based DTs have been explored for wireless channel modeling and calibration \cite{ruah2024calibrating,pegurri2025toward}. DTs are well-suited to generating context-specific synthetic data on demand: given the current operating regime (e.g., slice type, load level), a network DT can generate plausible nominal samples for calibration without requiring measurement collection.

However, DT-generated data is subject to a distribution mismatch relative to real observations, arising from model inaccuracies, environment changes, or unmodeled interference. Naively using DT-generated data as real calibration data in conformal inference invalidates the resulting p-values and FDR guarantees \cite{bashari2025synthetic,bashari2025statistical}. The use of synthetic data in conformal inference has been studied in general offline settings \cite{bashari2025synthetic,bashari2025statistical}, but its extension to online anomaly detection in telecommunications has not been addressed. Synthetic data has been used to augment training sets and synthesize rare anomalies \cite{rivera2020anomaly,kim2024enhancing,chen2025generative,wagenstetter2024synthetically}, but not to support calibrated, guarantee-preserving online detection.

\subsection{Novelty}

Table~\ref{tab:comparison} summarizes the key properties of representative existing methods compared to C-PP-COAD. Overall, no existing anomaly detection method controls FDR in a formal, distribution-free sense, while enabling the use of synthetic data and incorporating context into their anomaly statistics.

\begin{table}[t]
\caption{Comparison of Anomaly Detection Methods}
\label{tab:comparison}
\setlength{\tabcolsep}{3pt}
\small
\begin{tabular}{lccc}
\toprule
\textbf{Method} & \textbf{FDR control} & \textbf{Synth.\ data} & \textbf{Context-aware} \\
\midrule
Threshold/EWMA \cite{wang2018anomaly} & \texttimes & \texttimes & \texttimes \\
Autoencoder/LSTM \cite{zhou2017anomaly,hundman2018detecting} & \texttimes & \texttimes & \texttimes \\
GNN-based O-RAN \cite{shami2025ran} & \texttimes & \texttimes & \checkmark \\
COAD \cite{rebjock2021online} & \checkmark & \texttimes & \texttimes \\
\textbf{C-PP-COAD (proposed)} & \checkmark & \checkmark & \checkmark \\
\bottomrule
\end{tabular}
\end{table}

\section{PROBLEM DEFINITION}
\label{sec:problem}

In this section, we formally describe the setting and performance objectives for the context-aware online anomaly detection problem that we consider in this paper.

\subsection{SETTING}

We study an online monitoring system operating across discrete time steps $t = 1, 2, \ldots$ At each time $t$, the system observes a data point $X_t$ along with contextual information $C_t$. The \textit{context} $C_t$ is categorical, taking values in a discrete set $\mathcal{C}$. Each context value $C_t = c \in \mathcal{C}$ defines a specific data-generating mechanism, and we denote the nominal distribution for a given context $C_t = c$ as $P({X|C=c})$. These distributions are typically unknown and may exhibit significant variability across contexts. No assumption is made on the sequence of contexts $\{C_t\}_{t\geq 1}$, which is treated as an individual sequence.

To provide some examples, in a healthcare setting, the data point $X_t$ could represent the lab test results for a patient at time $t$, and the context $C_t$ might indicate demographic or clinical subgroups (e.g., age group) \cite{hauskrecht2013outlier}. In an industrial monitoring system, the data point $X_t$ may correspond to sensor readings, and the context $C_t$ could encode the operational mode of a machine (e.g., idle, active, calibration phase) \cite{farshchi2018contextual}. Finally, in a telecommunications network the data point $X_t$ may collect key performance indicators (KPIs) from the protocol interfaces, while the context $C_t$ gathers information about the network state in terms of, e.g., traffic load and resource utilization \cite{mahrez2023benchmarking,parwez2017big,li2019anomaly}.

The objective of anomaly detection is to assess whether each data point $X_t$ conforms to the expected distribution $P(X|C_t)$ of its associated context $C_t$. Specifically, we seek to determine whether $X_t$ is typical, i.e., an \textit{inlier}, under the distribution $P(X|C_t)$, or whether it significantly deviates from this distribution, in which case it is flagged as an \textit{anomaly}.

The problem of assessing whether a newly observed test point $X_t$ conforms to the context-specific distribution $P({X|C_t})$ can be formalized as a \textit{test} with the null \textit{hypothesis}
\begin{equation}
\label{eq:hypothesis}
\mathcal{H}_t: X_t \sim P({X|C_t}).
\end{equation}
Accordingly, \textit{rejecting} the null hypothesis $\mathcal{H}_t$ indicates that $X_t$ is an anomaly with respect to the distribution $P({X|C_t})$. We define as $A_t$ the indicator variable
\begin{equation}
    A_t = \begin{cases}
        0 \quad \text{if }X_t\text{ is an inlier, i.e., }X_t\sim P(X|C_t),\\
        1\quad \text{otherwise.}
    \end{cases}
\end{equation}

Our goal is to sequentially test the hypotheses $\mathcal{H}_t$, while controlling the proportion of \textit{false anomalies} (see Appendix~\ref{sec:background} for background on multiple hypothesis testing). To this end, we assume access to an \emph{arbitrary pre-trained} \textit{score function} $s(X|C)$, which quantifies how unusual the data point $X|C$ is as a sample from the distribution $P({X|C})$. Higher values of the score $s(X|C)$ indicate stronger evidence that $X|C$ does not conform to the distribution $P({X|C})$ \cite{bashari2025robust}. As discussed in Sec.~\ref{sec:intro}, many score functions have been introduced in the literature. For instance, supervised score functions may be obtained from binary classifiers trained to distinguish nominal from anomalous samples \cite{goldstein2016comparative,breiman2001random}, semi-supervised scores can be defined via the decision function of one-class classifiers trained on nominal data \cite{scholkopf2001estimating}, and unsupervised scores may quantify deviation from nominal structure through distances to cluster centroids \cite{lazarevic2005feature,lloyd1982least}.

To guarantee statistical performance requirements, for each time $t$, the system leverages a fresh batch of \textit{calibration data points} $\mathcal{D}_t = \{X_t^i\}_{i = 1}^n$ consisting of i.i.d.\ samples from distribution $P({X|C_t})$. These data points provide a baseline for inlier data that can be used to calibrate the score function $s(X|C)$. In practice, context-dependent data may be scarce, and thus it is preferable to use calibration data only when there is a high chance of a data point $X_t$ being anomalous.

To regulate the use of calibration data, we allow the monitor to first test each data point $X_t$ using \textit{synthetic calibration data} $\tilde{\mathcal{D}}_t = \{\tilde{X}_t^i\}_{i = 1}^{\tilde{n}}$. The synthetic dataset $\tilde{\mathcal{D}}_t$ consists of i.i.d.\ data points from a distribution that is generally distinct from the true data distribution $P(X|C)$. One approach to generating synthetic data is through \textit{digital twins}, virtual models that simulate real-world processes \cite{jones2020characterising}.

\begin{figure*}[t!]
    \centering
    \includegraphics[width=\textwidth]{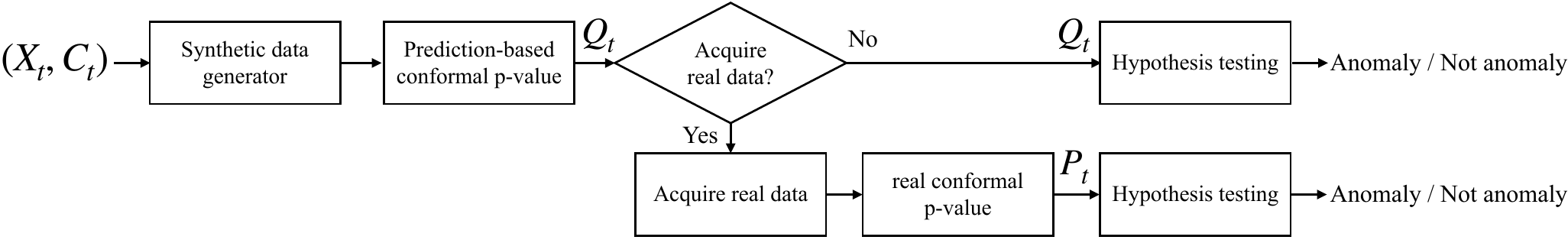}
    \caption{An overview of C-PP-COAD.}
    \label{fig:overview}
\end{figure*}

\subsection{PROBLEM DEFINITION}
\label{sec:metrics}

Using any pre-trained anomaly score $s(X|C)$, we are interested in designing an online anomaly detection framework that (\emph{i}) ensures that the fraction of false anomalies is no larger than a desired target $\alpha$, while making a best effort at (\emph{ii}) maximizing the fraction of true detected anomalies and (\emph{iii}) reducing the reliance on real calibration data.

A generic anomaly detection framework operates as follows:
\begin{enumerate}
   \item  \textit{Anomaly measure:} Given the current input $(X_t, C_t)$, the corresponding score $s(X_t|C_t)$, as well as the available synthetic and/or real calibration data, an anomaly measure $Z_t$ is computed.

    \item  \textit{Time-varying threshold:} A threshold $\alpha_t$ is applied to the anomaly measure $Z_t$ yielding the decision
     \begin{equation}
 \label{eq:anomaly_detection}
     \hat{A}_t =
     \begin{cases}
         0 \;\text{(no anomaly)} \quad &\text{if}\; Z_t > \alpha_t,\\
         1\; \text{(anomaly)}\quad &\text{if}\; Z_t \leq \alpha_t.
     \end{cases}
 \end{equation}
Accordingly, the binary variable $\hat{A}_t \in \{0,1\}$ is the estimate of the indicator $A_t$ produced by the system at time $t$.
\end{enumerate}

Restating the design goals, we thus wish to control the fraction of time instants $t$ in which there is no anomaly -- i.e., $A_t=0$ -- and yet the system detected an anomaly, $\hat{A}_t=1$, while also increasing the fraction of time instants $t$ in which the system detects a true anomaly -- i.e., $A_t=\hat{A}_t=1$.

To formalize this objective, let the weighted average of false anomalies be
\begin{equation}
\label{eq:Et}
    F_t = \sum_{\tau=1}^{t} \delta^{t-\tau} \hat{A}_{\tau}(1-A_\tau),
\end{equation}
with a \textit{decaying-memory factor} $0<\delta<1$. The weight $\delta^{t-\tau}$ in (\ref{eq:Et}) ensures that more recent false anomalies are given greater importance, while older false anomalies gradually diminish in influence. Define also the weighted average count of anomaly detections as
\begin{equation}
R_t = \sum_{\tau=1}^{t} \delta^{t-\tau} \hat{A}_\tau.
\end{equation}

With these definitions, the fraction of false anomalies is measured by the \textit{smoothed-decaying-memory} FDR (sFDR) \cite{rebjock2021online}
\begin{equation}
\label{eq:memory_decaying_FDR}
    \text{FDR}_t = \mathbb{E}_{C^t} \left[ \frac{F_t}{R_t +\eta} \right],
\end{equation}
where $\eta > 0$ is a small smoothing parameter and the average $\mathbb{E}_{C^t}[\cdot]$ in (\ref{eq:memory_decaying_FDR}) is evaluated with respect to the nominal distribution $\Pi_{\tau = 1}^tP(X_\tau|C_\tau)$, where $C^t = (C_1, \ldots, C_t)$ represents an \emph{arbitrary} context sequence.

The decaying-memory formulation in (\ref{eq:Et})--(\ref{eq:memory_decaying_FDR}) is motivated by a known failure mode of standard online FDR procedures: in the absence of anomalies, the LORD rejection thresholds $\alpha_t$ decrease monotonically until they collapse to near zero --- a phenomenon termed \emph{$\alpha$-death} \cite{rebjock2021online}. Once $\alpha_t \approx 0$, the detector becomes unable to flag any anomaly, even when they occur. The decaying memory prevents this by ensuring that long stretches of non-rejection do not accumulate unbounded weight in the FDR denominator.

The design goal is to ensure that the sFDR in (\ref{eq:memory_decaying_FDR}) is maintained below a target level $\alpha \in (0, 1)$ for all times $t$, i.e.,
\begin{equation}
\label{eq:fdr_condition}
    \text{FDR}_t \leq \alpha,
\end{equation}
while making a best effort at maximizing the number of detected true anomalies and minimizing the use of real calibration data. The inequality (\ref{eq:fdr_condition}) must hold irrespective of the quality of the underlying score function $s(X|C)$ and for any sequence of context variables.

The fraction of detected true anomalies and the rate of use of real data are measured by the following metrics:
\begin{itemize}
    \item \textbf{Power:} The power measures the system's ability to correctly identify true anomalies and is defined as
\begin{equation}
\label{eq:power}
\text{Power}_t = \mathbb{E}_{C^t}\left[\frac{\sum_{\tau = 1}^{t}\delta^{t-\tau} \hat{A}_\tau A_\tau}{\sum_{\tau = 1}^{t}\delta^{t-\tau}  A_\tau + \eta}\right].
\end{equation}
Accordingly, $\text{Power}_t$ measures the average proportion of anomalies that are successfully detected.

\item \textbf{Cumulative data acquisition rate:} Let $U_t \in \{0, 1\}$ be a binary variable indicating whether real data was used at time $t$. The cumulative data acquisition rate (CDAR) up to time $t$ is defined as the time-weighted average
\begin{equation}
\label{eq:cdar}
\text{CDAR}_t = \mathbb{E}_{C^t}\left[\sum_{\tau = 1}^{t}\delta^{t-\tau} U_\tau\right].
\end{equation}
The CDAR corresponds to the time-weighted average of the number of times that real-world calibration data is used, and serves as a proxy for operational cost.
\end{itemize}

\section{CONTEXT-AWARE ONLINE CONFORMAL ANOMALY DETECTION WITH PREDICTION-POWERED DATA ACQUISITION}
\label{sec:methodology}

We develop C-PP-COAD in three incremental steps. We start from the real-data baseline (COAD), motivate its cost, then introduce proxy p-values from a digital twin, yielding PO-COAD, which is fast but loses validity, and finally combine the two via active p-values and context-aware acquisition to obtain C-PP-COAD, which is both valid and data-efficient. Fig.~\ref{fig:overview} shows the resulting framework.

\subsection{THE REAL-DATA BASELINE (COAD)}
\label{sec:coad_baseline}

A natural starting point is to build p-values exclusively from real calibration data. At each time $t$, a fresh batch of nominal samples $\mathcal{D}_t = \{X_t^i\}_{i=1}^{n}$ matching context $C_t$ yields the conformal p-value
\begin{equation}
\label{eq:real_pval}
P_t = \frac{\sum_{i=1}^{n} \mathbb{I}[s(X_t^i | C_t) \geq s(X_t | C_t)]}{n + 1},
\end{equation}
which measures the rank of the test score among calibration scores: a small $P_t$ indicates that $X_t$ is more anomalous than most nominal samples. Setting the anomaly measure $Z_t = P_t$ and running the LORD online FDR procedure yields \emph{COAD} \cite{rebjock2021online}, which satisfies the sFDR constraint (\ref{eq:fdr_condition}) for all $t$. The cost is that real data must be acquired at \emph{every} time step, driving the CDAR to its maximum value of $1/(1-\delta)$.

\subsection{PROXY P-VALUES FROM SYNTHETIC DATA}
\label{sec:data_ac}

The cost of COAD motivates using DT-generated data instead of real calibration data. Given a synthetic calibration dataset $\tilde{\mathcal{D}}_t = \{\tilde{X}_t^i\}_{i=1}^{\tilde{n}}$ drawn from the DT for context $C_t$, the \emph{proxy conformal p-value} is
\begin{equation}
\label{eq:proxy_pval}
    Q_t = \frac{\sum_{i = 1}^{\tilde{n}}\mathbb{I}[s(\tilde{X}_t^i|C_t)\geq s(X_t|C_t)]}{\tilde{n} + 1},
\end{equation}
where $\mathbb{I}[\cdot]$ denotes the indicator function. A small $Q_t$ indicates that $X_t$ scores higher than most synthetic nominal samples, providing evidence of an anomaly.

Setting $Z_t = Q_t$ directly yields \emph{PO-COAD}, which incurs zero real data acquisition cost. However, because DT samples are drawn from an approximation of the true nominal distribution, $Q_t$ is \emph{not} a valid conformal p-value: it does not satisfy $\Pr[Q_t \leq u \mid \mathcal{H}_t] \leq u$ in general \cite{bashari2025synthetic,bashari2025statistical}. Plugging an invalid p-value into LORD violates the sFDR constraint (\ref{eq:fdr_condition}), as confirmed experimentally in Sec.~\ref{sec:experiments}. The proxy $Q_t$ is nonetheless useful as a \emph{signal}, as it tells us \emph{when} acquiring real data is worth the cost.

\subsection{ACTIVE P-VALUES AND PP-COAD}

The active p-value framework of Xu et al.~\cite{xu2025active} provides a principled way to use the proxy $Q_t$ as a gating signal while retaining validity. Real data $\mathcal{D}_t$ is acquired with probability
\begin{equation}
\label{eq:p_real}
     p^{\text{real}}(Q_t) =1 - \gamma Q_t,
\end{equation}
where $\gamma \in (0,1]$ controls the acquisition propensity: a large $Q_t$ (observation looks normal to the DT) reduces the acquisition probability, while a small $Q_t$ (DT suspects an anomaly) triggers near-certain real data collection. Letting $U_t \in \{0,1\}$ denote whether real data is acquired,
\begin{equation}
\label{eq:should_query}
    U_t \mid Q_t \sim \text{Bern}\left(p^{\text{real}}(Q_t)\right),
\end{equation}
the \emph{active p-value} \cite{xu2025active} is defined as
\begin{equation}
\label{eq:active_pvalue}
    Z_t = (1 - U_t) \cdot Q_t + U_t \cdot (1-\gamma)^{-1} \cdot P_t,
\end{equation}
where $P_t$ is the real conformal p-value in (\ref{eq:real_pval}), computed only if $U_t = 1$. When real data is not acquired ($U_t = 0$), $Z_t = Q_t$ is used directly; when it is ($U_t = 1$), $Z_t$ becomes an inflated version of the valid p-value $P_t$. Xu et al.\ show that $Z_t$ is super-uniform under $\mathcal{H}_t$ for any choice of $\gamma$, so FDR control is preserved.

Using a fixed scalar $\gamma$ (context-agnostic version) yields \emph{PP-COAD}. This already improves over COAD by reducing real data cost, at the expense of a power penalty proportional to the inflation factor $(1-\gamma)^{-1}$.

In order to apply the detection rule (\ref{eq:anomaly_detection}), we select the threshold $\alpha_t$ via the LORD algorithm \cite{rebjock2021online}, which updates the threshold at time step $t$ as
\begin{equation}
\label{eq:lord_threshold}
    \alpha_t = \alpha \eta\tilde{\zeta_t} + \alpha \sum_j \delta^{t- \rho_j}\zeta_{t-j},
\end{equation}
where $\rho_j = \min\{t \geq 0 \mid \sum_{i = 1}^t \hat{A}_i \geq j\}$ denotes the time of the $j$th anomaly detection decision ($\rho_j = \infty$ if no such decision has occurred); we defined $\tilde{\zeta_t} = \max(\zeta_t, 1-\delta)$; and $\{\zeta_t\}_{t = 1}^\infty$ is a non-increasing sequence summing to 1. A conventional choice is $\zeta_t \propto \log(\min(t,2))/(t\exp{\sqrt{\log t}})$, scaled to sum to 1 \cite{ramdas2017online}.

Alternative procedures for determining $\alpha_t$ to satisfy requirement (\ref{eq:fdr_condition}) include SAFFRON \cite{ramdas2018saffron} and ADDIS \cite{tian2019addis}.

\subsection{C-PP-COAD: CONTEXT-AWARE ACQUISITION}
\label{sec:cppoad_full}

The scalar $\gamma$ in PP-COAD treats all contexts identically. In practice, the DT's fidelity varies by operating regime: a DT trained predominantly on User Datagram Protocol (UDP) traffic may poorly represent the flow-level statistics of rarer Internet Control Message Protocol (ICMP) traffic, leading to unreliable proxy p-values in that context. C-PP-COAD replaces $\gamma$ with a \emph{context-dependent} parameter $\gamma(C_t) \in (0,1]$, extending (\ref{eq:p_real})--(\ref{eq:active_pvalue}) by substituting $\gamma \leftarrow \gamma(C_t)$:
\begin{equation}
\label{eq:p_real_ctx}
     p^{\text{real}}(Q_t, C_t) =1 - \gamma(C_t) Q_t.
\end{equation}
A higher $\gamma(C_t)$ in contexts where the DT is reliable reduces real data cost; a lower $\gamma(C_t)$ in contexts where the DT is unreliable increases the chance of querying real data. Crucially, statistical validity holds regardless of the choice of $\gamma(C_t)$ \cite{xu2025active}.

As further discussed in Appendix~\ref{sec:choosing_gamma}, $\gamma(C_t)$ can be estimated from held-out nominal validation data by measuring the one-sided Kolmogorov-Smirnov deviation of the proxy p-values $\{Q_t\}$ from uniformity. The complete C-PP-COAD procedure is depicted in Fig.~\ref{fig:overview}.

\subsection{THEORETICAL GUARANTEES}

C-PP-COAD ensures the following theoretical guarantee, whose proof can be found in Appendix~\ref{app:proof}.

\begin{proposition}
\label{prop:prop}
Fix any score function $s(X|C)$. For any context sequence $\{C_t\}_{t\geq 1}$, assuming the sequence $\{X_t\}_{t\geq 1}$ is i.i.d.\ conditioned on $\{C_t\}_{t\geq 1}$, C-PP-COAD guarantees control of the sFDR (\ref{eq:memory_decaying_FDR}) at the target level $\alpha$, i.e.,
\[
\text{FDR}_t \leq \alpha, \quad \text{for all time steps } t \geq 1.
\]
\end{proposition}

\subsection{THE C-PP-COAD FAMILY}
\label{sec:other_benchmarks}

The incremental development above yields a family of methods that trade off data acquisition cost, detection power, and statistical validity. Table~\ref{tab:method_family} summarizes the design choices. COAD (\textbf{Step 1}) and C-COAD (the context-aware variant of COAD) always query real data and are guaranteed to satisfy (\ref{eq:fdr_condition}), but incur maximum CDAR. PO-COAD and C-PO-COAD (\textbf{Step 2}) use only proxy p-values and acquire no real data, but lack formal FDR guarantees. PP-COAD (\textbf{Step 3}) and the full C-PP-COAD combine both signals via active p-values, reducing CDAR while preserving the sFDR guarantee.

\begin{table}[t]
\caption{The C-PP-COAD Family of Methods}
\label{tab:method_family}
\setlength{\tabcolsep}{4pt}
\begin{tabular}{lcccc}
\toprule
\textbf{Method} & \textbf{Context} & \textbf{Synth.} & \textbf{Real} & \textbf{sFDR} \\
\midrule
COAD \cite{rebjock2021online}   & \texttimes & \texttimes & always & \checkmark \\
C-COAD                          & \checkmark & \texttimes & always & \checkmark \\
PO-COAD                         & \texttimes & always     & \texttimes & \texttimes \\
C-PO-COAD                       & \checkmark & always     & \texttimes & \texttimes \\
PP-COAD                         & \texttimes & gating     & adaptive & \checkmark \\
\textbf{C-PP-COAD (proposed)}   & \checkmark & gating     & adaptive & \checkmark \\
\bottomrule
\end{tabular}
\end{table}

\section{EXPERIMENTS}
\label{sec:experiments}

In this section, we evaluate C-PP-COAD on four datasets spanning healthcare and telecommunications: thyroid disease detection \cite{asuncion2007uci}, synthetic O-RAN conflict detection \cite{shami2025ran}, 5G network intrusion detection \cite{samarakoon2022nidd}, and O-RAN UE throughput degradation detection \cite{polese2022coloran}. In all experiments, we control the sFDR below the target $\alpha = 0.1$, a standard choice in the FDR literature \cite{benjamini1995controlling,ramdas2017online}.

The main focus of the evaluation is on \emph{context-aware} methods. The first three experiments (Thyroid, O-RAN, and Thyroid classifier comparison) report results for context-aware benchmarks only. The 5G-NIDD experiment additionally includes a context-agnostic comparison to illustrate why accounting for protocol context improves FDR control and detection power. A key observation is that proxy-only methods cannot guarantee sFDR control: PO-COAD fails in the context-agnostic 5G-NIDD comparison, and C-PO-COAD fails in every context-aware experiment (Thyroid, O-RAN, ColO-RAN), because proxy p-values depend on the fidelity of the digital twin model. C-PP-COAD detects poor proxy quality via the active p-value gating mechanism and always maintains the sFDR guarantee.

\subsection{BENCHMARKS}
\label{sec:exp_benchmarks}

We evaluate C-PP-COAD against all benchmarks introduced in Sec.~\ref{sec:other_benchmarks}, along with two additional baselines. The first is \emph{Stat-AD} (Static Anomaly Detector), a fixed-threshold detector that declares an anomaly when $s(X_t \mid C_t) > s_\alpha(C_t)$, where $s_\alpha(C_t)$ is the $(1-\alpha)$-quantile of the score estimated on training data. The second is \emph{FC-COAD} (Full-Conformal Online Anomaly Detector) \cite{lee2025full}, the method of Lee et al.\ (2025), which computes a real conformal p-value $P_t$ at each step and rejects whenever $P_t \leq \alpha$, without any sequential FDR bookkeeping. Because it applies a fixed per-step threshold rather than an adaptive sequential procedure, FC-COAD does not control the sFDR over time.

All methods train an anomaly score function $s(X \mid C)$. PP and PO variants additionally train a synthetic data generator (digital twin, DT), while all methods except PO ones use fresh real calibration data during online testing. We adopt a unified data split across experiments: one third of the data is used to train the score function, one third to train the DT (when applicable), and the remainder is used for online calibration, evenly partitioned across time steps. PO methods instead use this final portion solely for DT training. The DT is trained offline; at test time, querying it amounts to drawing samples from a fitted model, which is computationally negligible relative to data acquisition costs.

The DT models the context-dependent data distribution via a Gaussian Mixture Model (GMM), $\text{GMM}(X \mid C)$, trained on the DT split. GMMs are appropriate for the moderate-dimensional, categorical-context datasets considered here; more expressive generators (e.g., VAEs, normalizing flows) can be substituted without affecting the validity guarantees of Proposition~\ref{prop:prop}.

We consider three representative anomaly scores: a random forest classifier~\cite{breiman2001random} (supervised), a one-class SVM with an RBF kernel~\cite{scholkopf2001estimating} (semi-supervised), and a k-means distance-based detector~\cite{lloyd1982least,lazarevic2005feature} (unsupervised). Context-aware methods train separate scores per context, whereas context-agnostic methods use a single global score.

\subsection{DETECTION OF THYROID DYSFUNCTIONS}
\label{sec:thyroid}

\subsubsection{Task Description}

We conduct the first experiment on the Thyroid disease dataset from the UCI Machine Learning Repository \cite{asuncion2007uci}. The dataset contains 29 numerical medical features for 7,200 patient records, with labels indicating normal thyroid function or thyroid dysfunction (hypothyroid and hyperthyroid). Following standard anomaly detection benchmarking practice \cite{goldstein2016comparative}, we treat thyroid dysfunction cases as anomalies and normal patient records as the nominal class. The anomaly rate is approximately 7\%. We define the context $C_t$ based on age group by partitioning the age feature into two bins ($0\text{--}50$ and $50+$), resulting in two distinct contexts.

Since the dataset contains missing values, we employ an imputation procedure following Appendix~\ref{sec:missing_values}: missing values in continuous features are replaced by the training-set median, and categorical features by the mode.

\subsubsection{On the Impact of the Score Function}

\begin{figure*}[!t]
\centering
\includegraphics[width=0.9\textwidth]{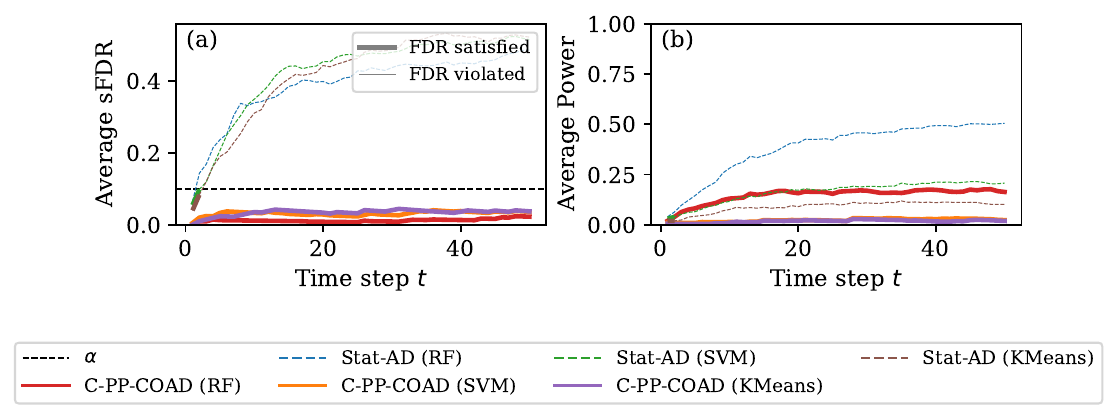}
\caption{Performance of supervised (random forest), unsupervised (clustering), and semi-supervised (one-class SVM) score functions on the Thyroid disease dataset, evaluated using a conventional fixed-threshold strategy and C-PP-COAD. The two panels show the average decaying memory sFDR (\ref{eq:memory_decaying_FDR}) and average power (\ref{eq:power}), respectively, as a function of time during testing ($T=50$ steps). Thin lines indicate violations of the sFDR guarantee.}
\label{fig:classifier}
\end{figure*}

As the first experiment, we showcase the benefits of applying C-PP-COAD to the three different score functions discussed in Sec.~\ref{sec:exp_benchmarks} as compared to using a conventional fixed-threshold approach. We ran the experiment 100 times over random data splits, evaluating over $T = 50$ time steps. We set the memory decay rate to $\delta = 0.95$ and the score tuning parameter in (\ref{eq:gamma_c}) to $\lambda = 5$.

The results, illustrated in Fig.~\ref{fig:classifier}, demonstrate that while the raw classification methods often yield higher detection power, they tend to violate the desired FDR constraint. In contrast, applying C-PP-COAD consistently bounds the FDR below $\alpha = 0.1$, albeit at the cost of reduced power. Thin lines in the plot indicate FDR violations, while thick lines represent cases where the FDR constraint is satisfied. The figure also demonstrates the advantages of supervised methods over alternative unsupervised and semi-supervised techniques.

\subsubsection{Comparison with Benchmarks}
\label{sec:thyroid_benchmarks}

\begin{figure*}[!t]
\centering
\includegraphics[width=0.9\textwidth]{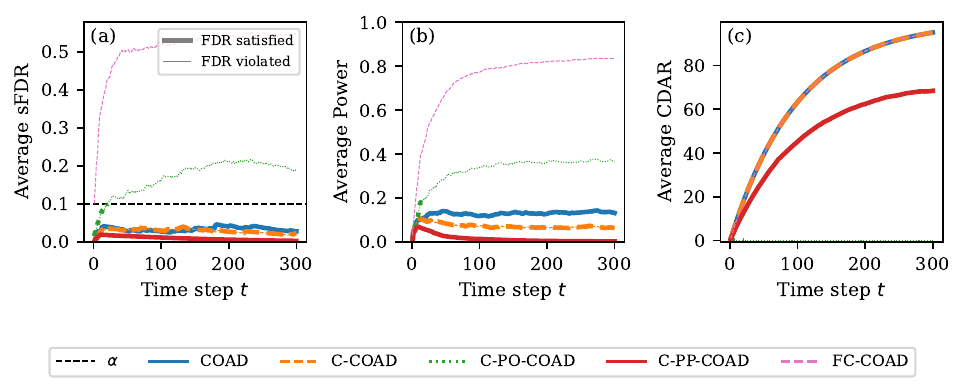}
\caption{Performance of C-PP-COAD and context-aware benchmark methods on the Thyroid disease dataset ($T=300$ steps). The three panels show the sFDR (\ref{eq:memory_decaying_FDR}), average power (\ref{eq:power}), and average CDAR (\ref{eq:cdar}), respectively, as a function of time during testing. Thin lines indicate violations of the sFDR guarantee.}
\label{fig:thyroid_context}
\end{figure*}

We use the supervised classifier from Sec.~\ref{sec:exp_benchmarks} as the score function and compare all context-aware benchmark methods introduced in Sec.~\ref{sec:exp_benchmarks}. Performance is evaluated in terms of sFDR, power, and CDAR, averaged over 100 runs with random data splits over $T = 300$ time steps. We set the memory decay rate to $\delta = 0.99$ and the score tuning parameter to $\lambda = 5$.

The left panel shows the average decaying-memory sFDR. In line with Proposition~\ref{prop:prop}, COAD, C-COAD, and C-PP-COAD satisfy the target sFDR constraint of $\alpha = 0.1$. C-PO-COAD violates it because its proxy p-values are not valid conformal p-values, as using synthetic data only does not provide the super-uniformity guarantee under $\mathcal{H}_0$, leading to spurious detections. FC-COAD also violates the constraint, since its fixed per-step threshold $\alpha$ does not adapt to the history of past decisions, causing false discoveries to accumulate over time. Thick lines indicate satisfaction of the sFDR constraint and thin lines indicate violations.

The middle panel reports average power. The methods with the highest power, i.e., Stat-AD and FC-COAD, both violate the sFDR constraint, and their elevated power comes at the cost of FDR control. Among methods that maintain the guarantee, C-PP-COAD achieves lower power than COAD in this dataset, as the active p-value gate is triggered conservatively when the proxy quality estimate $\hat{\gamma}(C)$ is low, causing C-PP-COAD to skip some detections to preserve validity.

The right panel displays the CDAR. C-PO-COAD and Stat-AD incur zero real data acquisition cost (their curves coincide with the bottom of the right panel at CDAR = 0). FC-COAD queries real data at every step, matching COAD's acquisition cost. C-PP-COAD reduces data acquisition relative to COAD and C-COAD while preserving sFDR control.

\subsection{CONFLICT DETECTION IN O-RAN}
\label{sec:oran}

\subsubsection{Task Description}

We evaluate C-PP-COAD on a synthetic conflict detection task for O-RAN, following the KPI monitoring framework of \cite{shami2025ran}. The goal is to detect xApp-induced KPI degradations, which are treated as anomalies.

Each data sample $X_t \in \mathbb{R}^{20}$ consists of 10 continuous KPI measurements (e.g., throughput, latency, signal-to-noise ratio), 5 xApp control signal values, and 5 channel quality indicators. Under normal operation, the network exhibits two regimes: a \emph{stable} regime (70\% of nominal samples) with KPI measurements drawn from $\mathcal{N}(0,1)$, and a \emph{high-load} regime (30\% of nominal samples) with KPI measurements drawn from $\mathcal{N}(4.0,1.5)$, reflecting elevated-but-valid traffic conditions not fully captured by the digital twin. Anomalies represent xApp-induced KPI degradations with KPI measurements drawn from $\mathcal{N}(6.5,1.5)$, which partially overlaps the high-load normal regime.

We generate 10,000 samples and define the context variable $C \in \{0,1,2,3\}$ as the quartile of the channel quality index, which is independent of the stable/high-load split. Approximately $10\%$ of samples are anomalies. Additional details on data construction are provided in Appendix~\ref{sec:app_oran}.

Unless stated otherwise, we set the memory decay factor to $\delta = 0.95$ and the data acquisition parameter to $\lambda = 5$.

\subsubsection{Comparison with Benchmarks}

\begin{figure*}[!t]
\centering
\includegraphics[width=0.9\textwidth]{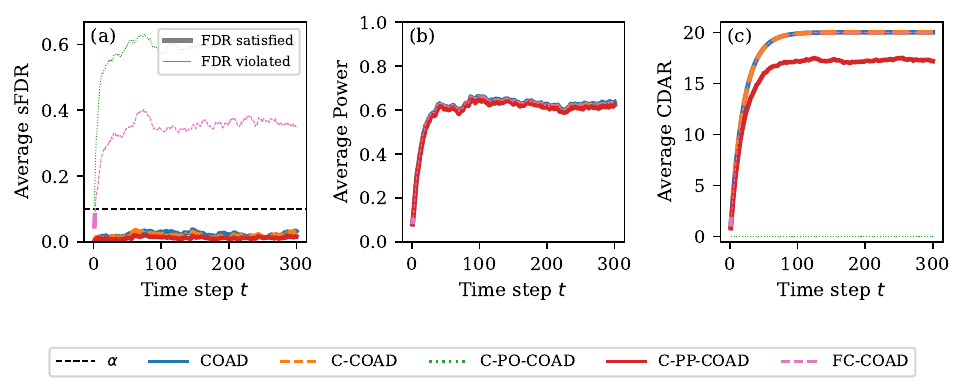}
\caption{Performance of COAD, C-PP-COAD, and context-aware benchmark methods on the O-RAN conflict detection dataset ($T=300$ steps). The three panels show the sFDR (\ref{eq:memory_decaying_FDR}), average power (\ref{eq:power}), and average CDAR (\ref{eq:cdar}), respectively, as a function of time during testing. In the first two panels, thin lines indicate violations of the sFDR guarantee.}
\label{fig:combined2}
\end{figure*}

In this experiment, we assess the performance of COAD, C-PP-COAD, and the context-aware benchmarks discussed in Sec.~\ref{sec:exp_benchmarks}. As in Sec.~\ref{sec:thyroid_benchmarks}, all three performance metrics (sFDR, power, and CDAR) are averaged over 100 independent runs with random data splits over $T = 300$ time steps.

The left panel shows the average decaying-memory sFDR. In line with Proposition~\ref{prop:prop}, COAD, C-COAD, and C-PP-COAD satisfy the target sFDR constraint of $\alpha = 0.1$. C-PO-COAD violates it because its proxy p-values are not valid conformal p-values, as using synthetic data only does not provide the super-uniformity guarantee under $\mathcal{H}_0$. Specifically, the per-context GMMs are trained on lower-scoring nominal samples and do not generate high-load normal samples, so $Q_t \approx 0$ for high-load normals, causing spurious detections. FC-COAD also fails to control the sFDR, since it applies a fixed threshold $\alpha$ at each step without accounting for past decisions, leading false discoveries to accumulate over time. Thick lines indicate satisfaction of the sFDR constraint and thin lines indicate violations.

The middle panel reports average power. All methods achieve similar average power in this experiment, as the overlapping normal and anomaly distributions make detection challenging regardless of the method. The methods that violate the sFDR constraint (C-PO-COAD, Stat-AD, FC-COAD) do not gain meaningfully higher power than C-PP-COAD, confirming that the sFDR violation is not justified by a power benefit.

The right panel displays the CDAR. C-PO-COAD and Stat-AD incur zero real data acquisition cost (their curves coincide with the bottom of the right panel at CDAR = 0). FC-COAD queries real data at every step. C-PP-COAD reduces data acquisition relative to COAD and C-COAD while preserving sFDR control.

The impact of the data acquisition parameter $\lambda$ on detection power and data efficiency is further investigated via an ablation study reported in Appendix~\ref{app:exp_ablation}.

\subsection{5G NETWORK INTRUSION DETECTION}
\label{sec:5gnidd}

\subsubsection{Task Description}

We now evaluate C-PP-COAD on network intrusion detection using the 5G-NIDD dataset \cite{samarakoon2022nidd}, a benchmark of real 5G network traffic flows collected over a dedicated 5G testbed. The dataset includes 31 numerical flow-level features, e.g., duration, packet counts, byte rates, inter-arrival times, covering both benign and malicious traffic spanning denial-of-service (DoS), distributed denial-of-service (DDoS), and port scanning attacks. Malicious traffic flows are treated as anomalies and benign flows as nominal. Since malicious flows represent a small fraction of the original dataset, we subsample the benign traffic to achieve a balanced 10\% anomaly rate, yielding $N = 18{,}000$ samples. We define the context $C_t \in \{0,1,2,3\}$ as the transport protocol, i.e., UDP, TCP, ICMP, or other, yielding four distinct traffic profiles that differ substantially in flow duration, packet-rate patterns, and byte-rate distributions. We use a shallow random forest with depth 1 as the anomaly score function, reflecting a resource-efficient baseline appropriate for real-time inference in network elements. We set $\delta = 0.95$ and $\lambda = 5$.

\subsubsection{Comparison with Benchmarks}

\begin{figure*}[!t]
\centering
\includegraphics[width=0.9\textwidth]{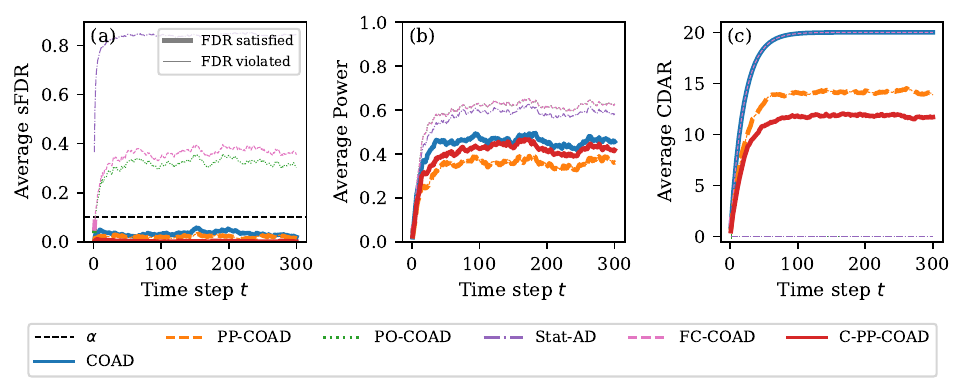}
\caption{Performance of C-PP-COAD and context-agnostic benchmark methods on the 5G-NIDD network intrusion detection dataset ($T=300$ steps, $\delta=0.95$). The three panels show the average sFDR (\ref{eq:memory_decaying_FDR}), average power (\ref{eq:power}), and average CDAR (\ref{eq:cdar}), respectively, as a function of time. Thin lines indicate violations of the sFDR guarantee.}
\label{fig:5gnidd_nocontext}
\end{figure*}

Performance is evaluated in terms of sFDR, power, and CDAR, averaged over 100 runs with random data splits over $T = 300$ time steps. Results are shown in Fig.~\ref{fig:5gnidd_nocontext}.

The left panel shows the average decaying-memory sFDR. Thick lines indicate satisfaction of the sFDR constraint and thin lines indicate violations. In line with Proposition~\ref{prop:prop}, COAD, PP-COAD, and C-PP-COAD satisfy the target sFDR constraint of $\alpha = 0.1$. PO-COAD violates it because its proxy p-values are not valid conformal p-values. The distribution mismatch between the DT and the true nominal distribution means $Q_t$ is not super-uniform under $\mathcal{H}_0$, leading to spurious detections. FC-COAD also violates the sFDR constraint, since rejecting whenever $P_t \leq \alpha$ without adjusting for the number of past rejections causes false discoveries to accumulate over time.

The middle panel reports average power. C-PP-COAD achieves detection power comparable to COAD. The methods with the highest average power, i.e., Stat-AD, PO-COAD, and FC-COAD, all violate the sFDR constraint, and thus their elevated power comes at the cost of FDR control. Note that PO-COAD and FC-COAD achieve identical power in this experiment, as both effectively reject every anomaly due to near-zero p-values at anomalous observations.

The right panel displays the CDAR. PO-COAD and Stat-AD incur zero real data acquisition cost. COAD queries real data at every time step. C-PP-COAD substantially reduces real data acquisition relative to COAD, demonstrating that synthetic data is used at a significant fraction of time steps.

\subsection{UE THROUGHPUT DEGRADATION DETECTION}
\label{sec:coloran}

\subsubsection{Task Description}

We finally evaluate C-PP-COAD on radio access network (RAN) monitoring using the ColO-RAN dataset \cite{polese2022coloran}, collected from the Colosseum large-scale wireless network emulator. Each sample $X_t$ contains per-UE measurements including downlink/uplink throughput, signal-to-noise ratio (SNR), modulation and coding scheme (MCS), block error rate (BLER), reference signal received power (RSRP), and path loss, recorded under three O-RAN scheduling policies: Round-Robin (RR), Proportional Fair (PF), and Maximum Throughput (MaxThroughput). We define anomalies as UE throughput below the 10th percentile of the scheduling-policy-specific downlink bitrate distribution, modeling service level agreement (SLA) violations due to resource starvation or radio degradation. The context $C_t \in \{0,1,2\}$ encodes the scheduling policy. We use $N = 48{,}000$ samples with a 10\% anomaly rate, a random forest with depth 3 as the score function, and set $\delta = 0.95$ and $\lambda = 5$.

The per-scheduling-policy digital twins are trained on nominal UE measurements recorded in Colosseum emulator runs where no resource starvation or radio degradation events were introduced, representing purely normal network behavior. This introduces a coverage gap: UEs with throughput just above the anomaly threshold (i.e., marginally normal UEs) are rare in such controlled runs and thus underrepresented in the per-policy Gaussian DT. Because the DT concentrates its probability mass on UEs with clearly normal throughput, it assigns negligible likelihood to these \emph{borderline-normal} UEs. As a consequence, the proxy p-value $Q_t$ for such a UE is near zero (the same value that would indicate an anomaly), even though the UE is fully nominal (i.e., $A_t = 0$). Under C-PO-COAD, which relies solely on $Q_t$ for detection, these UEs are spuriously flagged as anomalous, violating the sFDR constraint.

\subsubsection{Comparison with Benchmarks}

\begin{figure*}[!t]
\centering
\includegraphics[width=0.9\textwidth]{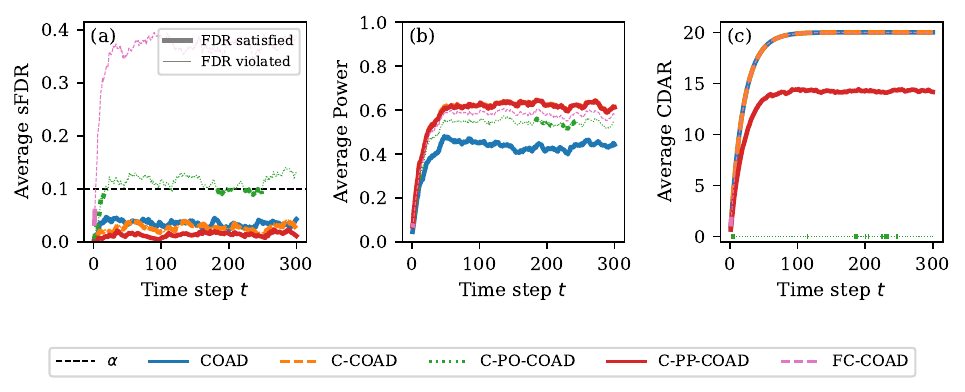}
\caption{Performance of C-PP-COAD and context-aware benchmark methods on the ColO-RAN UE throughput degradation dataset ($T=300$ steps, $\delta=0.95$). The three panels show the average sFDR (\ref{eq:memory_decaying_FDR}), average power (\ref{eq:power}), and average CDAR (\ref{eq:cdar}), respectively, as a function of time. Thin lines indicate violations of the sFDR guarantee.}
\label{fig:coloran_context}
\end{figure*}

Fig.~\ref{fig:coloran_context} reports the three performance metrics, namely sFDR, power, and CDAR, averaged over 100 independent runs with random data splits over $T = 300$ time steps. The left panel shows the average decaying-memory sFDR. As expected from Proposition~\ref{prop:prop}, COAD, C-COAD, and C-PP-COAD remain below the target level $\alpha = 0.1$. C-PO-COAD fails to satisfy this constraint, as its per-context proxy p-values are not valid conformal p-values. FC-COAD also violates the sFDR constraint, since a fixed per-step threshold $\alpha$ applied without reference to past decisions causes false discoveries to build up over time.

The middle panel reports average power. C-PP-COAD achieves the highest detection power among all methods, closely followed by C-COAD. Notably, FC-COAD, despite violating the sFDR constraint, does not gain a power advantage over C-PP-COAD or C-COAD, confirming that the FDR violation does not translate into improved detection in this experiment. COAD records the lowest power, as its context-agnostic calibration fails to exploit the scheduling-policy-dependent score structure.

The right panel displays the CDAR. COAD, C-COAD, and FC-COAD all incur maximum real data acquisition cost, querying real calibration data at every time step. C-PO-COAD incurs zero acquisition cost, relying entirely on synthetic data. C-PP-COAD substantially reduces data acquisition relative to COAD and C-COAD while preserving the sFDR guarantee, demonstrating that the proxy gating mechanism effectively identifies time steps where synthetic data alone suffices.

\section{CONCLUSION}
\label{sec:conclusion}

We proposed C-PP-COAD, a prediction-powered framework for online anomaly detection that improves data efficiency while providing rigorous false discovery rate guarantees. By adaptively combining synthetic and real calibration data via conformal inference and active p-values, C-PP-COAD reduces reliance on costly real-world observations while remaining compatible with arbitrary pre-trained anomaly scores and robust to missing features. Experiments on healthcare and real-world telecommunication datasets, including 5G network intrusion detection and O-RAN UE throughput monitoring, showed that C-PP-COAD maintains sFDR control and achieves higher detection power than conventional conformal methods, with substantially reduced real data acquisition cost.

Future work can explore adaptive assessment of synthetic data quality across contexts and further improvements in the power--efficiency tradeoff.

\appendices

\section{Reproducibility}
\label{app:code}

The codebase for the experiments in this paper is publicly available at \url{https://github.com/amirfar76/c-pp-coad}.

\section{Background on Multiple Hypothesis Testing}
\label{sec:background}

\subsection{Hypothesis Testing}
Hypothesis testing is a fundamental statistical tool used to assess whether observed data provides sufficient evidence to reject a given assumption, known as the null hypothesis $\mathcal{H}_0$. A standard hypothesis test begins with the specification of a null hypothesis $\mathcal{H}_0$, which asserts that there is no significant deviation from a specified model. A test statistic, denoted as $T(X)$, is then computed based on the observed data $X$. The probability distribution of $T(X)$ under $\mathcal{H}_0$ is used to evaluate the extremity of the observed value.

To assess evidence against $\mathcal{H}_0$, a p-value is commonly used. A valid p-value $p$ satisfies the condition
\begin{equation}
\mathrm{Pr}[p\leq u|\mathcal{H}_0] \leq u \quad \text{for all} \; u\in[0,1],
\end{equation}
ensuring that under $\mathcal{H}_0$, the probability of observing a p-value at most $u$ does not exceed $u$ \cite{rice2007mathematical}. If $p \leq \alpha$, the null hypothesis $\mathcal{H}_0$ is rejected; otherwise, there is insufficient evidence to reject it. This framework ensures that the probability of incorrectly rejecting $\mathcal{H}_0$ (a Type I error) does not exceed $\alpha$.

\subsection{Multiple Hypothesis Testing}

In modern applications, particularly in high-dimensional statistics and machine learning, multiple hypothesis tests are often conducted simultaneously. Testing them independently using their respective p-values increases the risk of false discoveries, necessitating additional control mechanisms.

A key metric in this setting is the \textit{false discovery rate} (FDR), which quantifies the expected proportion of false discoveries among all rejected hypotheses. Formally, let $R$ denote the total number of rejected hypotheses and $V$ the number of false discoveries. The FDR is then defined as
\begin{equation}
\label{eq:FDR}
\text{FDR} = \mathbb{E} \left[ \frac{V}{\max(R,1)} \right].
\end{equation}

To control the FDR at a predefined level $\alpha$, simple thresholding of individual p-values at $\alpha$ is insufficient. Instead, FDR-controlling procedures such as the Benjamini--Hochberg (BH) procedure \cite{benjamini1995controlling} adjust the rejection thresholds to ensure that the expected FDR remains below $\alpha$.

Other FDR-controlling procedures include the Benjamini--Yekutieli procedure \cite{benjamini2001control}, which is more conservative and valid under arbitrary dependence among p-values, and adaptive procedures such as Storey's method \cite{storey2002direct}, which estimate the proportion of true null hypotheses to tighten rejection thresholds. For sequential or online settings, procedures like LORD \cite{javanmard2018online}, SAFFRON \cite{ramdas2018saffron}, and ADDIS \cite{tian2019addis} extend FDR control to cases where hypotheses arrive over time and decisions must be made without access to future data.

\section{Proof of Proposition~\ref{prop:prop}}
\label{app:proof}

\begin{IEEEproof}
The proof relies on two key facts:
\begin{enumerate}
    \item[\textit{(i)}] \emph{Validity of $Z_t$.} The test statistic $Z_t$ is a valid (superuniform) p-value under the null hypothesis $\mathcal{H}_t$, i.e., $\Pr[Z_t \leq u \mid \mathcal{H}_t] \leq u$ for all $u \in [0,1]$. This follows from \cite{xu2025active}: validity of the active p-value requires only that the real conformal p-value $P_t$ be valid under $\mathcal{H}_t$, which in turn follows from the exchangeability of the real calibration data conditioned on $C_t$. Crucially, this guarantee holds regardless of the quality of the synthetic data or the choice of $\gamma(C_t)$.

    \item[\textit{(ii)}] \emph{Independence of $\{Z_t\}_{t \geq 1}$.} The p-values $Z_t$ are independent across time steps, since fresh calibration datasets are drawn at each step and the test points are i.i.d.\ conditioned on the context sequence.

    \item[\textit{(iii)}] \emph{sFDR control via LORD.} Given a sequence of independent valid p-values, the LORD algorithm is guaranteed to control the sFDR at level $\alpha$ for all $t \geq 1$, as proven in \cite{rebjock2021online}.
\end{enumerate}

Combining (i)--(iii) immediately yields $\text{FDR}_t \leq \alpha$ for all $t \geq 1$.
\end{IEEEproof}

\section{C-PP-COAD Algorithm}

\begin{algorithm}[h]
\caption{C-PP-COAD}
\label{alg:C-PP-COAD}
\begin{algorithmic}[1]
\FOR{each time step $t$}
    \STATE Observe test point $(X_t, C_t)$.
    \STATE Compute anomaly score $s(X_t | C_t)$.
    \STATE Generate synthetic calibration dataset $\tilde{\mathcal{D}}_t = \{\tilde{X}^i_t\}_{i = 1}^{\tilde{n}}$.
    \STATE Compute proxy p-value $Q_t$ using (\ref{eq:proxy_pval}).
    \STATE Sample indicator $U_t \sim \text{Bern}\left(p^{\text{real}}(Q_t,C_t)\right)$.
    \IF{$U_t = 1$}
        \STATE Collect real calibration data $\mathcal{D}_t = \{X^i_t\}_{i = 1}^{n}$.
        \STATE Compute the real p-value $P_t$ using (\ref{eq:real_pval}).
    \ENDIF
    \STATE Compute active p-value $Z_t$ using (\ref{eq:active_pvalue}).
    \STATE Compute adaptive rejection threshold $\alpha_t$ using (\ref{eq:lord_threshold}).
    \IF{$Z_t \leq \alpha_t$}
        \STATE Reject $\mathcal{H}_t$: classify $X_t$ as an anomaly.
    \ELSE
        \STATE Accept $\mathcal{H}_t$: classify $X_t$ as normal.
    \ENDIF
\ENDFOR
\end{algorithmic}
\end{algorithm}

\section{Optimizing and Extending C-PP-COAD}
\label{sec:optimizing}

In this section, we first propose a method for designing the context-specific parameters $\gamma(C)$ in the data acquisition probability (\ref{eq:p_real}), and then provide an extension of C-PP-COAD that handles missing values in the input data.

\subsection{Context-Aware Data Acquisition Probability}
\label{sec:choosing_gamma}

As discussed in Sec.~\ref{sec:data_ac}, a lower value of $\gamma(C)$ in (\ref{eq:p_real}) increases the probability of querying the true p-value $P$ in the active p-value (\ref{eq:active_pvalue}). Consequently, $\gamma(C)$ should ideally be higher when we have greater confidence in the synthetic data generator for context $C$, and lower when trust is more limited. More formally, $\gamma(C)$ should increase with the quality of the proxy p-values $Q$ produced by the synthetic data generator for context $C$.

To assess this quality, let $\mathcal{S}_C = \{\tilde{X}^i\}_{i = 1}^{|\mathcal{S}_C|}$ denote a set of synthetic data points generated by the simulator for context $C$, and let $\mathcal{V}_C = \{X^i\}_{i = 1}^{|\mathcal{V}_C|}$ represent a held-out set of inlier validation points from context $C$. For each data point $X^i \in \mathcal{V}_C$, we compute the statistic
\begin{equation}
\label{eq:test_pvals}
    p^i = \frac{\sum_{j = 1}^{|\mathcal{S}_C|} \mathbb{I}[s(\tilde{X}^j | C) \geq s(X^i | C)]}{|\mathcal{S}_C| + 1}.
\end{equation}

If $p^i$ is a valid p-value for the null hypothesis $\mathcal{H}^i$ that $X^i$ is an inlier, the superuniformity condition $\mathrm{Pr}[p_{\mathcal{V}_C}^i \leq x] \leq x$ for all $x \in [0,1]$ must hold. To assess this, we construct the empirical CDF
\begin{equation}
    \hat{F}_C(p) = \frac{1}{|\mathcal{V}_C|} \sum_{i = 1}^{|\mathcal{V}_C|} \mathbb{I}[p^i \leq p],
\end{equation}
and define the deviation metric
\begin{equation}
    D(C) = \sup_{0\leq p \leq 1} (\hat{F}_C(p) - p).
\end{equation}
The function $D(C)$ is closely related to the Kolmogorov--Smirnov (KS) distance~\cite{massey1951kolmogorov}, but captures only the positive deviation from the uniform CDF, which is the relevant direction for testing superuniformity. If $D(C) \leq 0$, the superuniformity condition is estimated to hold; if $D(C) > 0$, the proxy p-values may not be reliable.

Following this discussion, $\gamma(C)$ can be chosen as any decreasing function of $D(C)$ with range $(0,1]$. In our study, we use
\begin{equation}
\label{eq:gamma_c}
    \gamma(C) = \exp\left(-\lambda \max\left(0, D(C)\right)\right),
\end{equation}
where $\lambda > 0$ is a tuning parameter adjusting the weight of the score $D(C)$. Further discussion on this choice can be found in Sec.~\ref{app:exp_ablation}.

\subsection{Incomplete Observations}
\label{sec:missing_values}

To model a more realistic deployment scenario, we account for the possibility of missing values in the input vectors. Following \cite{zaffran2023conformal}, each data point is given by $X_t^{\text{obs}} = X_t \odot M_t \in \mathbb{R}^d$, where $M_t \in \{0,1\}^d$ is a binary mask indicating the presence or absence of features of the original data $X_t$, and $\odot$ is element-wise multiplication. A value of 1 at position $k$ in $M_t$ signifies that the $k$th element is missing.

We assume that the masks $M_t$ are independent of the data $X_t$ and are drawn i.i.d.\ across both calibration and test sets. This corresponds to the \textit{missing completely at random} (MCAR) setting from \cite{zaffran2023conformal}.

To handle missing values at test time, we adopt an \emph{impute-then-predict} approach~\cite{zaffran2023conformal}. We apply a fixed, pretrained imputation function $\Phi(X_t^{\text{obs}})$ that fills in the missing entries based on observed components. The imputed data points $\hat{X}_t = \Phi(X_t^{\text{obs}})$ are then used for calibration and prediction. Since the missingness mechanism is MCAR, the data points remain i.i.d., so the guarantees of Proposition~\ref{prop:prop} apply also in the presence of incomplete observations.

\section{Additional Experiments}
\label{sec:app_experiments}

\subsection{Ablation Study on the Parameter $\lambda$}
\label{app:exp_ablation}

We study the impact of $\lambda$ in (\ref{eq:gamma_c}), which controls the frequency of real calibration data acquisition in C-PP-COAD. Larger values of $\lambda$ increase the probability of querying real data, thereby trading data efficiency for potential gains in detection power.

We evaluate the average detection power and CDAR of C-PP-COAD for four linearly spaced values of $\lambda$ in $[1,10]$. Each configuration is repeated over 100 independent runs with random data splits. For each run, we report the final values of power and CDAR at the end of the testing horizon. Experiments are conducted for two target sFDR levels, $\alpha \in \{0.1, 0.2\}$.

\begin{figure}[h!]
\centering
\includegraphics[width=\columnwidth]{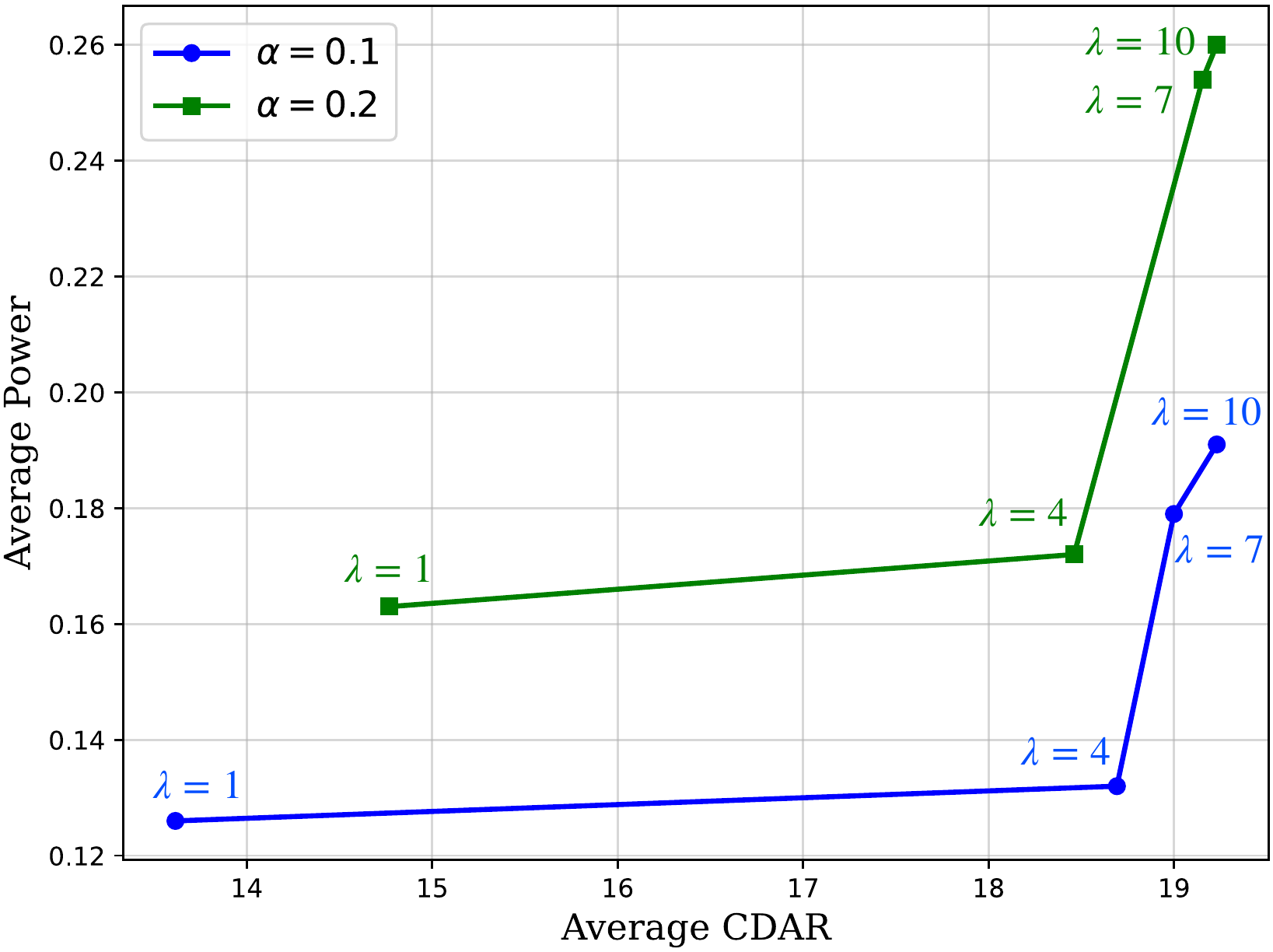}
\caption{Average CDAR and power achieved by C-PP-COAD for different values of the parameter $\lambda$ in (\ref{eq:gamma_c}), for two sFDR targets $\alpha \in \{0.1, 0.2\}$.}
\label{fig:lambda}
\end{figure}

The results are shown in Fig.~\ref{fig:lambda}. As $\lambda$ increases, detection power improves while data efficiency decreases, reflecting more frequent acquisition of real calibration data. This highlights a clear and tunable tradeoff between statistical power and reliance on real data. Importantly, the sFDR guarantee is preserved for all values of $\lambda$. Additionally, relaxing the target FDR level (larger $\alpha$) yields higher achievable power across all settings.

\section{Synthetic O-RAN Dataset Construction}
\label{sec:app_oran}

The synthetic O-RAN dataset models a multi-cell network where concurrent xApp control actions may induce KPI degradations. Each sample $X_t \in \mathbb{R}^{20}$ contains 10 continuous KPI measurements (features 1--10, e.g., throughput, latency, signal-to-noise ratio), 5 xApp control signal values (features 11--15), and 5 channel quality indicators (features 16--20).

Nominal samples follow a bimodal distribution reflecting two typical operating regimes:
\begin{itemize}
  \item \emph{Stable operation} (70\% of nominal): KPI measurements $\sim \mathcal{N}(0, 1)$; xApp signals and channel quality indicators $\sim \mathcal{N}(0, 1)$.
  \item \emph{High-load normal} (30\% of nominal): KPI measurements $\sim \mathcal{N}(4.0, 1.5)$, representing elevated-but-valid traffic conditions; xApp signals and channel quality indicators $\sim \mathcal{N}(0, 1)$.
\end{itemize}
Anomalous samples represent xApp-induced KPI degradations: KPI measurements $\sim \mathcal{N}(6.5, 1.5)$, which partially overlaps the high-load normal regime. xApp signals and channel quality indicators are drawn from $\mathcal{N}(0,1)$ for anomalous samples as well, so the score function must rely on the KPI features to distinguish anomalies.

The context variable $C \in \{0,1,2,3\}$ is defined as the quartile of the channel quality index (feature 16), computed from the nominal training data. Since channel quality is drawn from $\mathcal{N}(0,1)$ independently in both stable and high-load regimes, both regimes appear in every context. The digital twin (a single-component GMM) is fitted on the lower-scoring 70\% of nominal training samples within each context, which corresponds approximately to the stable-regime samples. Consequently, the per-context GMM does not generate high-load normal samples, creating a deliberate mismatch that causes C-PO-COAD to violate the sFDR constraint, while C-PP-COAD detects this mismatch via the active p-value gate and falls back to real calibration data.

\section*{ACKNOWLEDGMENT}
The work of A.~Farzaneh and O.~Simeone was supported by the European Research Council (ERC) under the European Union's Horizon Europe Programme (grant agreement No.\ 101198347). The work of O.~Simeone was also supported by an EPSRC Open Fellowship (EP/W024101/1) and by the EPSRC project (EP/X011852/1). The authors declare no competing interests.

\bibliographystyle{IEEEtran}
\bibliography{bib}

\begin{IEEEbiographynophoto}
{AMIRMOHAMMAD FARZANEH} (Member, IEEE) is currently a Postdoctoral Research Associate with the Institute for Intelligent Networked Systems (INSI), Northeastern University London, London, UK, and a Lecturer in Engineering Science with Oriel College and in Computer Science with Pembroke College, University of Oxford, Oxford, UK. Previously, he was a Postdoctoral Research Associate with King's College London, London, UK, from 2024 to 2026. He received his D.Phil.\ degree in Engineering Science from the University of Oxford, Oxford, UK, in 2024, and his B.Sc.\ degree in Electrical Engineering from Sharif University of Technology, Tehran, Iran. His research interests lie at the intersection of statistical machine learning, information theory, and network science, with a focus on developing rigorous reliability and uncertainty quantification guarantees for AI-driven systems in communication networks.
\end{IEEEbiographynophoto}

\begin{IEEEbiographynophoto}
{OSVALDO SIMEONE} (Fellow, IEEE) is currently a Professor of information engineering with Northeastern University London, where he co-directs the Institute for Intelligent Networked Systems (INSI), and a Visiting Professor with Aalborg University. Previously, he was with King's College London from 2017 to 2025 and New Jersey Institute of Technology (NJIT) from 2006 to 2017. He is the author of the textbooks Machine Learning for Engineers and Classical and Quantum Information Theory (Cambridge University Press), four monographs, two edited books, and more than 200 research journals and magazine papers. He is a fellow of IET. Among other recognitions, he was a co-recipient of the 2022 IEEE Communications Society Outstanding Paper Award, the 2021 IEEE Vehicular Technology Society Jack Neubauer Memorial Award, the 2019 IEEE Communication Society Best Tutorial Paper Award, the 2018 IEEE Signal Processing Best Paper Award, and the 2015 IEEE Communication Society Best Tutorial Paper Award. He was awarded an Advanced Grant by European Research Council (ERC) in 2025, an Open Fellowship by EPSRC in 2022, and a Consolidator Grant by ERC in 2016. He was a Distinguished Lecturer of the IEEE Communications Society from 2021 to 2022 and the IEEE Information Theory Society from 2017 to 2018.
\end{IEEEbiographynophoto}

\balance

\end{document}